# UAV Thermal Imagery for Inert Ordnance Screening: Multi Campaign Dataset Development, Object Detection, and Practical Recommendations

Chad Melton, PhD[1]
Annabelle Kelton[1]

[1] Oak Ridge National Laboratory, Geospatial Science and Human Security Division, Oak Ridge, Tennessee, USA.

## Abstract

Unexploded ordnance (UXO) continues to restrict civilian access, agricultural activity, infrastructure recovery, and environmental remediation in contaminated areas around the world. This study created a multi campaign UAV thermal image data set of inert ordnance, developed a labeled image set from collected imagery, tested object detection models, and identified practical considerations for humanitarian mine action and demining applications. Data were collected during four field campaigns in Tennessee under summer and winter conditions using inert mines, munitions, and other ordnance placed in short grass, tall vegetation, gravel, mulch, rock, compost, and compacted surfaces. Thermal imagery was collected under flight altitutes of 33 m and 15 m. The final source inventory contained 5,855 thermal image label pairs, including 918 positive images and 4,937 background images. After retaining all positive images and downsampling background images, the 33 m dataset contained 420 training and 106 validation images, while the 15 m dataset contained 629 training and 157 validation images. YOLOV11l and RT-DETR-R50 algorithms were trained and evaluated to develop an automated candidate detection model. Each model retained its native training and evaluation procedures, and the results were interpreted within each model rather than as a controlled architecture comparison. Within the YOLOV11l model, the 15 m dataset produced precision of 0.925, recall of 0.833, F1 of 0.877, mAP50 of 0.905, and mAP50-95 of 0.664. Within the RT-DETR-R50 model, the 15 m dataset produced precision of 0.911, recall of 0.750, F1 of 0.823, mAP50 of 0.740, and mAP50-95 of 0.511, compared with precision of 0.891, recall of 0.699, F1 of 0.783, mAP50 of 0.644, and mAP50-95 of 0.316 for the 33 m dataset. Practical recommendations include collecting thermal and RGB imagery together, incorporating varied surfaces and background only imagery, considering periods following changes in solar exposure, balancing survey coverage against target pixel representation, calibrating models with representative local data, and retaining qualified human review. The intended use is screening and prioritization for follow on technical survey or EOD assessment, and not a standalone clearance.

Keywords: unexploded ordnance; inert ordnance; unmanned aerial vehicle; thermal infrared imagery; dataset development; image annotation; object detection; humanitarian mine action; demining; decision support

## Introduction

According to data accessed April 1, 2026, Ukraine remains one of the most heavily explosive ordnance contaminated countries in the world, with the National Mine Action Program *Demining Ukraine* reporting that up to 144,000 km² of territory are potentially contaminated and require survey and clearance [1]. This contamination includes mines and other explosive remnants of war and continues to constrain civilian access, agricultural use, infrastructure recovery, and broader socioeconomic activity [2,3]. Current response activities rely on established mine action approaches including non technical survey, technical survey, clearance, and explosive ordnance disposal, consistent with international mine action terminology and operational practice [4,5]. In this context, UAV based sensing, including UAV mounted thermal imaging, may provide a useful supplementary capability by supporting faster, safer detection and mapping of suspect hazards prior to ground intervention [6].

## Background

Initiatives to detect UXOs are traditionally reliant on ground based investigations such as the use of hand held or vehicle mounted sensors, robotic systems, or trained animals. However, this style of discovery presents issues involving safety [7]. The detonation of these devices during an investigation could lead to equipment damage, soil contamination, and/or the death or injury of civilians and personnel [8].

To lessen these risks, new methods of UXO detection have been created to survey these areas remotely. Methods include drone based magnetometry [9], far range hyperspectral imagery [10], drone based transient electromagnetic systems [11], aerial synthetic aperture radar [12], and drone mounted ground penetrating radar [13]. With these methods come limitations such as limited detectability of certain materials [9,14] and/or aerial flight restrictions [9,12,13]. For example, transient electromagnetic systems aim to detect only conductive materials [14] while magnetometry aims to detect only ferromagnetic materials [9].

Similarly, hyperspectral imaging exhibits inconsistent results when detecting UXOs with a range of differing materials and colors, resulting in different spectral signatures [10]. This restriction limits the ability to detect plastic UXOs, such as PFM-1 mines. In addition to material detection restrictions, UAV mounted synthetic aperture radar, ground penetrating radar, and magnetometry have difficulties maintaining data integrity on a drone due to the lack of stability or signal range [9,12,13].

Thermal infrared imaging offers a complementary method because the sensor records naturally emitted surface radiation and can reveal temperature differences caused by material properties, solar loading, and heat transfer [15]. Unlike magnetometry or electromagnetic induction, thermal imaging is not restricted to the direct detection of ferromagnetic or electrically conductive materials. However, thermal sensors primarily measure radiation emitted from the surface. Therefore, detection of buried ordnance depends on whether the object creates a measurable temperature anomaly at the ground surface. It is important to consider that this response will change with burial depth, substrate, vegetation, weather, and time of day.

Researchers [16] conducted a study of PFM-1 landmine detection and identification capabilities of a small low altitude UAV mounted thermal sensor to detect PFM-1 (e.g. small, light anti personnel mines encased in plastic) mines. Due to their plastic outer shell, they are unable to be accurately detected by methods such as magnetometry and EM (electromagnetic method). To solve this issue, the study's aim was to investigate the accuracy, time constraints, risks, and accessibility of a UAV mounted thermal imaging system to detect the presence, orientation, and possible ballistic overlap of PFM-1 mines.

To simulate the environment of minefields in Afghanistan, a site with lower grass and rubble in Chenango State Park was chosen for the investigation. To analyze how temperature affects how the landmines emit heat, the low altitude drone flights occurred intermittently during sunrise and sunset in late summer.

The team discovered that KSF-1 casings of PFM-1 landmines were easily detectable by the thermal sensor with a 100 percent detection rate and detection sensitivity of 77.88 percent. Unique qualities of the study such as study site and orientation detectability analysis granted vital information for future study of UXO detection. Nevertheless, narrow drone altitude requirements, restricted data collection times, and a small collection site pose limiting factors to the application of this methodology.

In the study by Bajić and Potočnik [17], three meter thermal images were used to create and test a computational method based on deep learning on convolutional neural networks focused on detecting UXO. The imagery was collected using a drone mounted thermal sensor in Bosnia and Herzegovina. These raw data were then semi automatically annotated to build an evaluation dataset of 808 thermal images.

Two sets of object detection models were created: one to detect UXOs based on their classification and another for general UXO detection. Next, 11 classes of UXOs were classified based on the landmine dimensions from the images. 640 of these images were randomly chosen to train the deep neural networks, which included versions of YOLOv5.

Overall, precision was higher than 98% for all the YOLOv5 versions. The high percentage of precision proved that the YOLOv5 deep neural network can be used to accurately detect and classify UXOs based on their thermal signature. Yet, the small dataset limits the efficacy of reproduction with larger datasets. Despite this, Bajić et al. [17] created a computational workflow that is precise, reproducible, and adequately detects unexploded ordnances.

A study conducted by Fardoulis et al. [18] investigated the success of using thermal remote imagery to detect plastic and metal UXOs. To create a realistic investigation, fieldwork occurred at legacy minefields in the Sahara Desert. Data collection occurred during the hours of sunset and sunrise to understand the effects of ambient air temperature change. Arrangement of the landmines was vital to this study to validate the landmine locations; therefore, traditional military landmine placement tactics were used.

Ultimately, the investigation revealed that drone mounted thermal remote sensing systems are useful in detecting anomalies from anti tank and anti personnel landmines at legacy minefields in arid to semi arid environments. In addition, the scientists discovered factors that affect performance such as flying height, physical properties of individual landmines, diurnal elements, burial depth, the sun, the wind, and possibly cloud cover. Some limiting factors of this study included the lack of testing in areas with vegetation and differing burial depths. However, this study presents a clear starting point for future research of buried UXO detection techniques.

A similar inspection was conducted by Coulibaly et al. [19], where aerial infrared thermography was used to detect artisanal (i.e., handmade mines which are commonly found in Sahelian countries) UXOs in Burkina Faso. These landmines are difficult to detect due to their casings made of wood, ceramic, or plastic.

The study consisted of using a drone and thermal sensor at a low altitude to map areas of interest along an unpaved road. Three missions collected 750 thermal images each that were grouped and matched by a photogrammetry tool. With these orthomosaics, the photogrammetry algorithm processed the thermal contrast of the image to identify differences in temperature and generate a report.

This study concluded that UAV mounted infrared thermography can be used to detect buried plastic, ceramic, and wood encased artisanal landmines during ideal times of day along unpaved roads in Burkina Faso. Of the three missions, the highest percentage of detectability was 75 percent. Wooden, ceramic, and plastic were detectable by the system, however, metal landmines were not. Also, the study determined that the efficacy of the system would be decreased with the presence of vegetation and other buried objects in the survey area. Yet, this investigation exhibits a unique mathematical model and study design that can be reproduced for future studies.

*Study Objective and Contributions*

Previous studies have shown that UAV thermal imagery can be used to detect thermal anomalies associated with mines and other ordnance. However, among the studies reviewed here, relatively few describe in a single study the full sequence of designing field collection campaigns, converting the resulting imagery into a labeled object detection dataset, evaluating its suitability for automated screening, and translating the findings into practical guidance for future users.

To address this gap, the objective of this study was to create and document a field collected UAV thermal image dataset of surface and partially obscured inert ordnance, develop a reproducible workflow for reviewing and labeling the imagery, evaluate the suitability of the resulting datasets for object detection, and identify practical considerations for humanitarian mine action and demining organizations. This work contributes in four specific ways. First, the study created a raw thermal and RGB image data set across four campaigns that included different seasons, vegetation conditions, surface materials, and flight altitudes. Second, the study presents a process for image quality control, bounding box annotation, background image selection, and dataset partitioning. Third, YOLOV11l and RT-DETR-R50 were trained and evaluated to determine whether the labeled imagery could support automated candidate detection. The models were not used to identify a universally superior architecture because they retained different training and evaluation procedures. Lastly, field observations and model development experience were used to develop practical recommendations for future thermal data collection, analyst review, and follow on technical investigation.

# Methodology

*Workflow Overview*

The study was organized around four linked stages: raw thermal and RGB image collection; manual review, quality control, and labeling; dataset construction and object detection; and development of practical recommendations for analyst supported screening. The detailed sequence was mission planning, and location scouting and assessment → environmental and temporal scheduling → UAV thermal and RGB image collection → image quality control → target annotation → dataset construction → model training and validation → analyst review → prioritization for follow on ground investigation.

The experiments presented in this study directly evaluated the field collection, annotation, dataset development, and image level object detection stages under controlled conditions. The practical recommendations were developed from field observations, manual image review, labeling experience, dataset construction, and model behavior. Geospatial grouping of repeated detections, false alarm testing with the realistic number of background images, and operational clearance validation remain important next steps rather than completed portions of this study (see Fig 1).

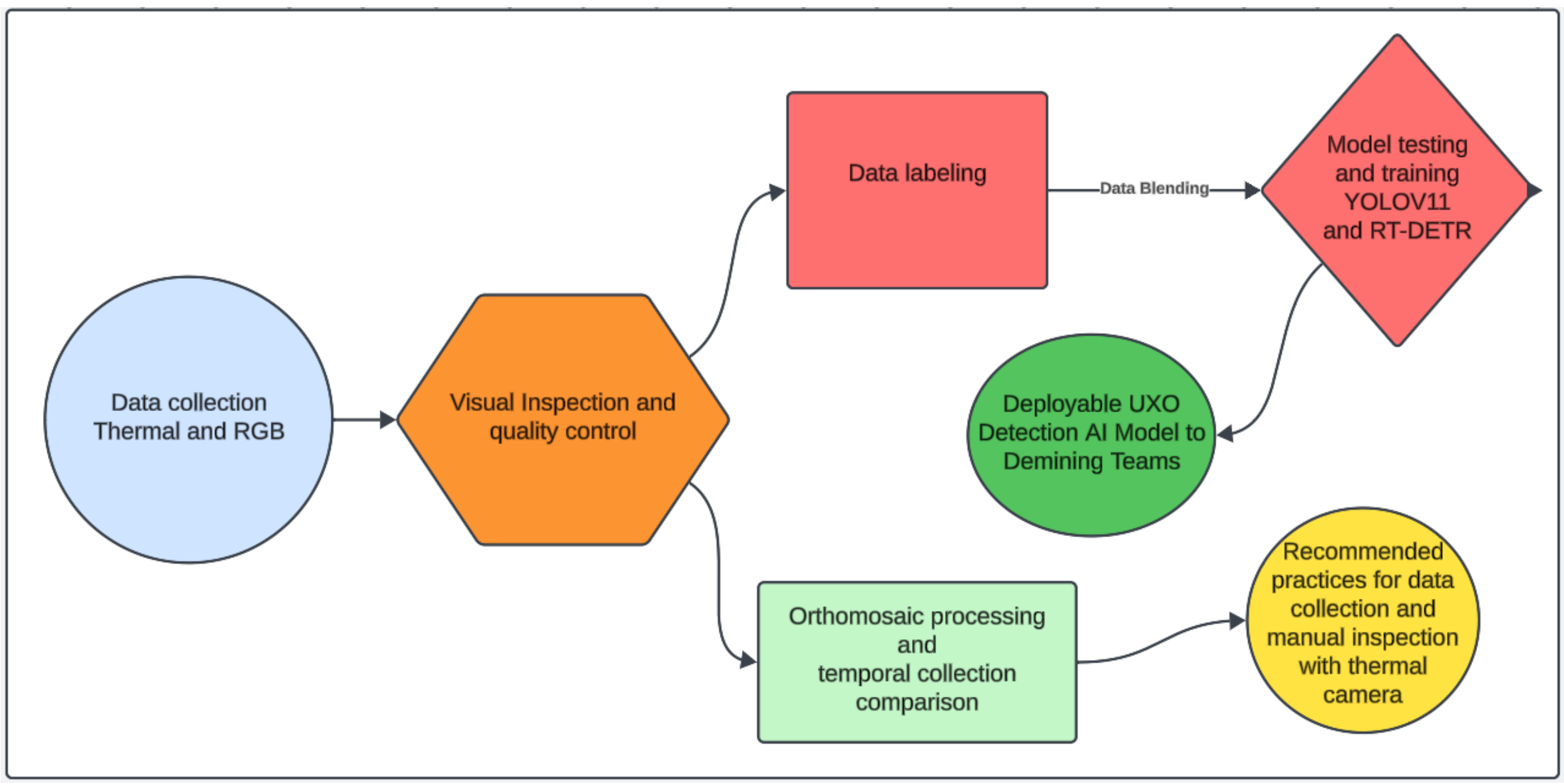

*Figure 1. Flowchart of presented methodology.*

*Field Campaigns and Experimental Design*

Data were collected during four field campaigns in Tennessee. Three campaigns were conducted at a flight altitude of 33 m, including two in Norris, TN and one in Georgetown, TN. The fourth campaign was conducted at approximately 15 m at the Norris location and at a landscaping supply site in Knoxville area. These campaigns were scheduled during late summer, winter, and early summer to capture differences in temperature, vegetation, surface materials, and solar loading.

During the two 33 m campaigns in Norris, 31 and 21 inert items were placed, respectively. Ordnance placement was conducted with guidance from EOD specialists from the U.S. Army 52nd EOD Group at Fort Campbell, Kentucky. Items were positioned in both tall and short vegetation, and the items placed in taller grass were marked with red survey flags. This design was intended to create a range of surface exposure and partial vegetation obstruction rather than reproduce the complete structure of an operational minefield.

For each two day campaign in Norris, flights began at approximately 15:00 on the first day and were conducted every hour until sunset. Data collection resumed at sunrise on the second day and continued every hour until approximately 15:00. Flights were programmed at approximately 33 m above ground level, producing an estimated ground sampling distance of approximately 2.6 cm per pixel. Front and side overlap were set between 80 and 85 percent to support orthomosaic development. Although the flight schedule did not capture one uninterrupted day, similar environmental conditions across the two collection days allowed the study to approximate a heating and cooling cycle. This schedule was necessary due to the availability of the EOD team specialist and inert ordnance.

The Georgetown campaign was conducted over approximately five hours and included five inert items. Three flights were completed during each hour, and the items were systematically moved between the south field, gravel road, and north field. This step was taken to increase variability in the surface material and background surrounding each target despite the smaller number of available items. The approximately 1.5 km² site consisted of grasses ranging from roughly 0.5 to 1 m in height and was divided near the center by a gravel road. Data were collected from approximately 10:00 to 14:20.

The fourth campaign was conducted in June 2026 at a landscaping supply business containing stockpiles of mulch, gravel, river rock, and compost, as well as access roads compacted by heavy machinery. This location was chosen because the variety of surface materials allowed the team to create several bare earth and low vegetation conditions. Inert items were placed on compacted gravel roads and selected landscape materials. Thermal and RGB data were collected with a Parrot ANAFI UKR at approximately 15 m above ground level during evening flights between 17:00 and 20:00.

Weather conditions varied substantially between seasons. The campaigns were scheduled with the goal of collecting imagery under a broad range of environmental conditions while still maintaining safe UAV flight operations (Table 1, Figure 2A-D).

*Table 1. Environmental conditions reported for each data collection campaign.*

| Collection | Site | Date | Sunrise | Sunset | Weather summary | Temperature range |
|---|---|---|---|---|---|---|
| 1 | Norris, TN | September 18, 2025 | 7:21 AM | 7:38 PM | Mostly clear and dry | 59°F to 85°F |
| 1 | Norris, TN | September 19, 2025 | 7:22 AM | 7:37 PM | Mostly clear, becoming somewhat cloudier later in the day | 59°F to 87°F |
| 2 | Georgetown, TN | December 9, 2025 | 7:36 AM | 5:27 PM | Clear and cold | 28°F to 55°F |
| 3 | Norris, TN | January 20, 2026 | 7:44 AM | 5:50 PM | Clear, cold, and very dry | 19°F to 58°F |
| 3 | Norris, TN | January 21, 2026 | 7:44 AM | 5:51 PM | Increasingly cloudy, with light rain in the evening | 18°F to 58°F |
| 4 | Norris, TN | June 16, 2026 | 6:18 AM | 8:55 PM | Mixed cloud cover, ranging from mostly cloudy to mostly clear; visibility generally 10 miles or greater | 61°F to 80°F |
| 4 | Norris, TN/Knoxville | June 17, 2026 | 6:18 AM | 8:55 PM | Warm summer conditions with variable cloud cover, including early overcast conditions and mostly clear periods; visibility generally 10 miles or greater | 61°F to 88°F |

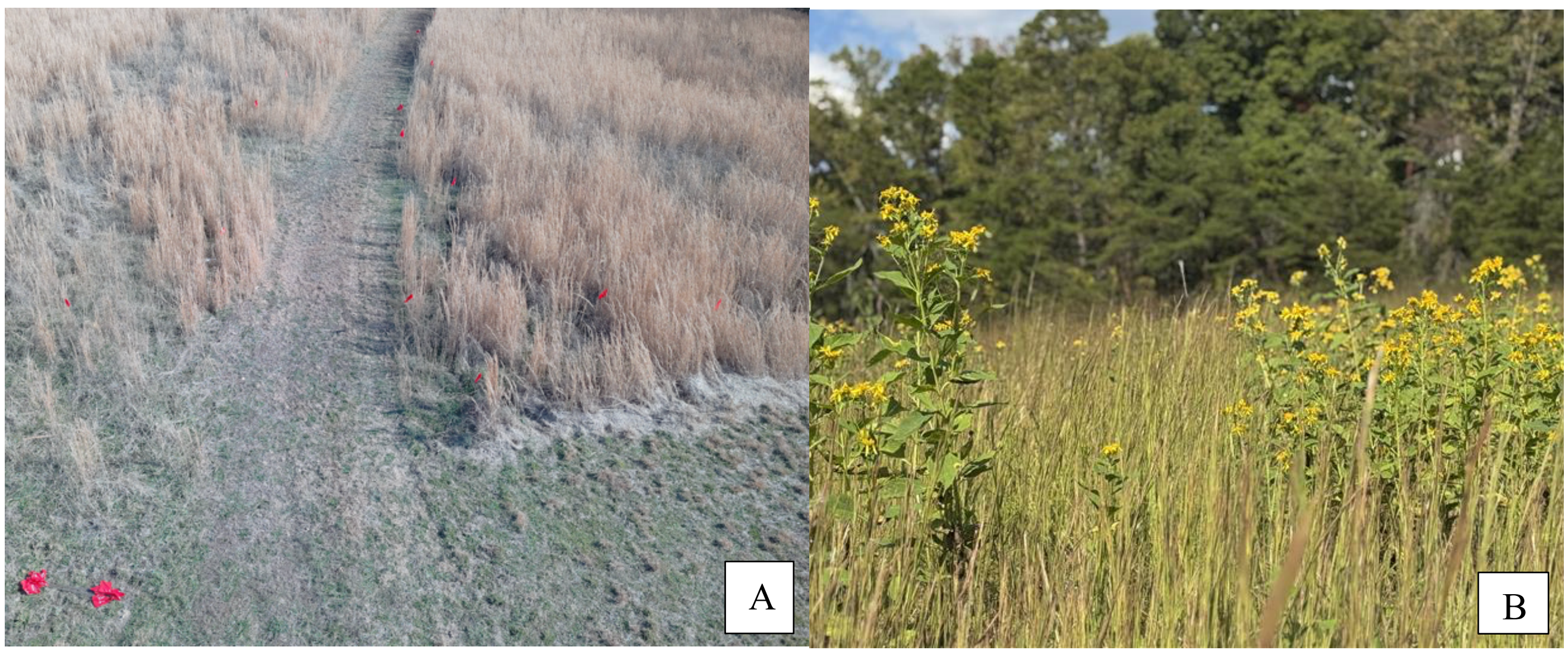

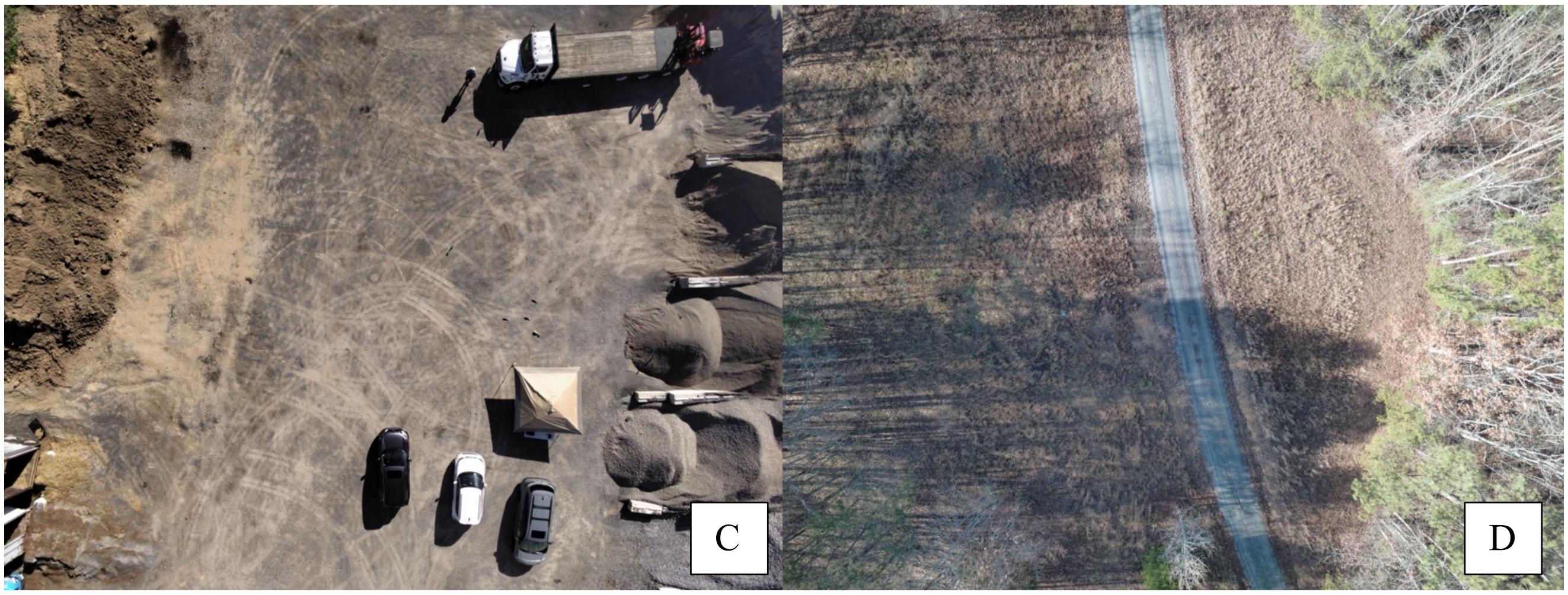

*Figure 2A-D. Image (A) Aerial image of the Norris, Tennessee, collection area showing recently cut and unmanaged field vegetation captured in January of 2026. Image (B) Image of typical field vegetation at the Norris site from September 2025. Image (C) displays the bare Earth site in Knoxville, and image (D) displays the collection site in Georgetown, TN.*

*Aircraft, Sensor Payloads, and Inert Ordnance*

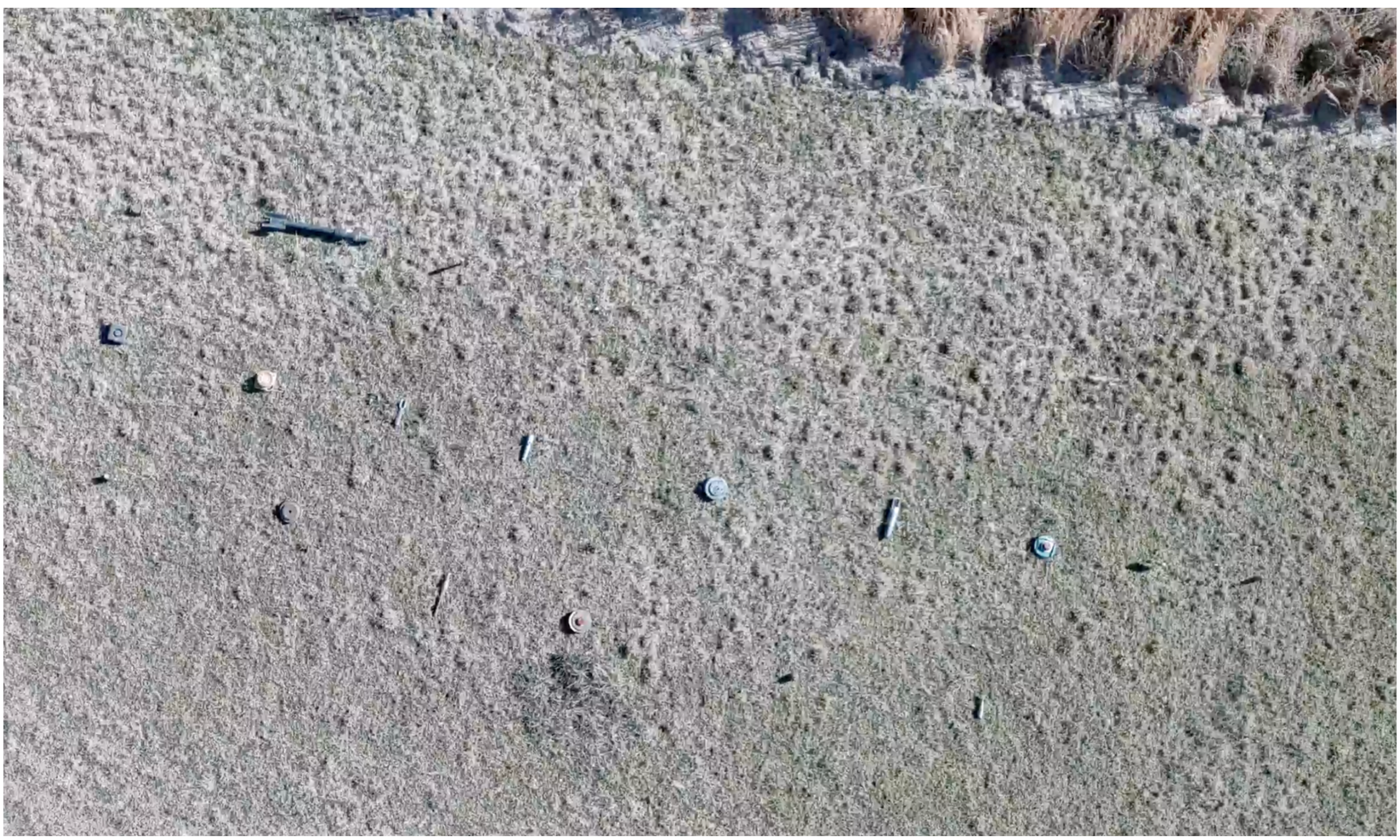

*Figure 3. Overhead image of inert ordnance training items arranged in recently cut field vegetation.*

*Aircraft and Sensor Payloads*

Flights were conducted with two UAV platforms, a Skydio X10 and a Parrot ANAFI UKR, which were used to collect thermal and RGB imagery [20,21].

The Skydio X10 sensor package combined high resolution RGB cameras with a radiometric thermal sensor. The visible spectrum payload included a 50.3 MP wide camera, a 64 MP narrow camera, and a 48 MP camera. The thermal payload consisted of a FLIR Boson+ uncooled VOx microbolometer with a 640 × 512 array and a reported thermal sensitivity below 30 mK NEDT. In combination, these sensors allowed thermal and high resolution visual imagery to be collected during the same flight [20].

*Inert Ordnance*

Inert ordnance was provided on loan by private collectors and the U.S. Army 52nd EOD Group at Fort Campbell, Kentucky, the Tennessee National Guard, and the John Sevier Hunter Education Center. The materials provided by private collectors included several U.S. manufactured mines, grenades, artillery components, and a Panzerfaust training item. The majority of the materials were supplied by the EOD group and included Soviet or Russian mines, artillery items, guided rockets, submunitions, and missiles. Overall, the training ordnance collection included anti personnel mines, anti tank mines, rockets, cluster munition components, hand grenades, directional fragmentation mines, mortar rounds, and tank artillery ejecta that ranging from WW2 through ongoing modern conflicts (see Figure 4A-F).

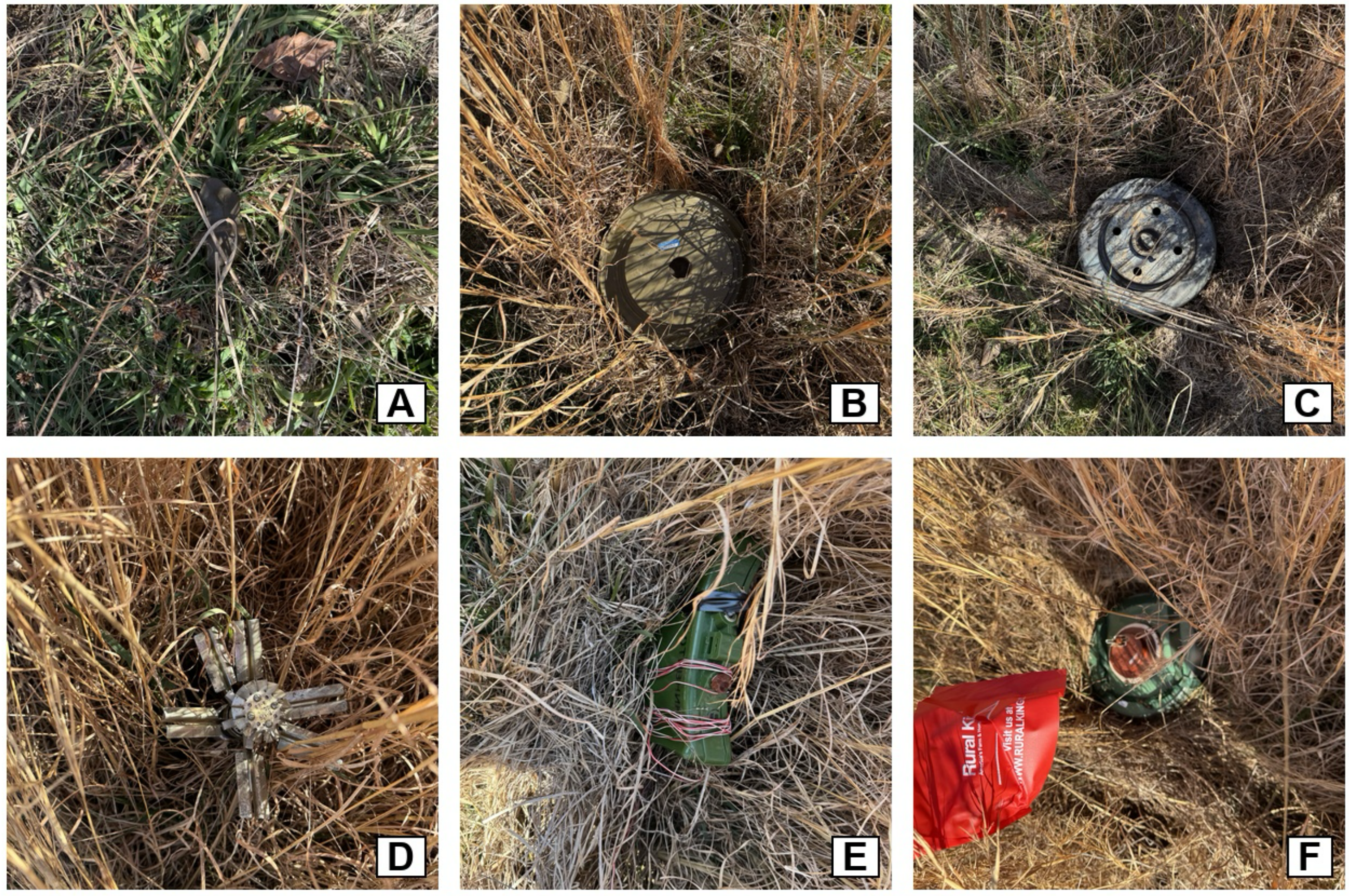

*Figure 4A-F. Inert ordnance placed in tall vegetation: (A) PFM-1; (B) TM-62M anti tank mine; (C) M6 anti tank mine; (D) 9N235 submunition; (E) M18A1 Claymore; and (F) TM-72 anti tank mine.*

The inert training items used throughout the four collection campaigns included:

Mon-50 LM, Gyata Apers LM, 3 Ptab 2.5 KO, Ptab 2.5, V-5KP1 rocket, OL-3.5 tandem rocket, TATG-9VM2 tandem, Recoilless rifle rocket, UKA-63 AT mine, Strela, s-5m, 9m14M guided missile, RPG-75, 60 mm mortar, M1 AT mine, Panzerfaust, M67 hand grenades, POMZ-2, PFM-1 Butterfly AP mine, TM-62 AT mines, MON-50 AP, tank artillery ejecta, VS-50 AT mine, TM-46 AT mine, TMA-3 mine, and 81mm mortar.

*Data Labeling and Dataset Preparation*

After each data collection campaign, the thermal imagery was manually reviewed and visible inert ordnance targets were labeled in CVAT [22] with bounding boxes. All of the original target categories were remapped to a single uxo class for object detection. Images containing at least one valid annotation were classified as positive, while images without a valid annotation were considered background candidates. It should be noted that ground truth was based on targets that could be identified in the images. Therefore, the evaluation measures the detection of annotated and image visible targets rather than every item known to have been placed at the site.

Background candidates substantially outnumbered positive images. To create a more manageable dataset for model development, all positive images were retained and the background candidates were randomly downsampled until the selected dataset contained approximately 70 percent positive images and 30 percent background images. A fixed random seed of 0 was used so that the sampling and partitioning could be reproduced. Positive and background images were then divided separately into an 80/20 training validation split and recombined to preserve approximately the same image composition in both subsets. This design was useful for model development and within dataset validation. However, it does not represent the low target prevalence that would likely occur during a real world survey.

For the 33 m analysis, labeled imagery from each of the high altitude campaigns was combined before model training. The finalized training set contained 313 images from the September 2025 collection, 855 from the December 9, 2025 collection, and 1,458 from the January 2026 collection, yielding 2,626 images. Of these, 368 contained at least one labeled UXO target and 2,258 were background images. All 368 positive images and 158 randomly selected background images were retained, producing a 526 image selected dataset. The training subset contained 420 images, including 294 positive and 126 background images with 619 annotated instances. The validation subset contained 106 images, including 74 positive and 32 background images with 176 annotated instances. Both object detection models used these training and validation partitions, although each retained its own training and evaluation procedures.

For the 15 m analysis from June 2026, the labeled imagery contained 3,229 image label pairs, including 550 positive images and 2,679 background images. All 550 positive images and 236 randomly selected background images were retained, producing a 786 image selected dataset. The training subset contained 629 images, including 440 positive and 189 background images with 810 annotated instances. The validation subset contained 157 images, including 110 positive and 47 background images with 192 annotated instances. The 15 m RT-DETR-R50 model was trained using these exact partitions so that both object detection models were based on the same training imagery and validation subset. The models were evaluated using their respective metrics (see Table 2).

*Table 2. Source image inventory used for dataset development before background resampling and train validation partitioning. The 33 m dataset combined three collections from September 2025, December 2025, and January 2026, while the 15 m dataset used the clean June 2026 collection.*

| Altitude | Flight Campaign | Total images | Positive images | Background images | Positive (%) | Background (%) |
|---|---|---|---|---|---|---|
| 33 m | September 2025 | 313 | 138 | 175 | 44.1 | 55.9 |
| 33 m | December 2025 Collection | 855 | 83 | 772 | 9.7 | 90.3 |
| 33 m | January Collection | 1,458 | 147 | 1,311 | 10.1 | 89.9 |
| 33 m Total | Combined dataset | 2,626 | 368 | 2,258 | 14 | 86 |
| 15 m | June 2026 Collection | 3,229 | 550 | 2,679 | 17 | 83 |
| Overall Total | All collections | 5,855 | 918 | 4,937 | 15.7 | 84.3 |

*YOLOV11l Object Detection*

For the YOLOV11, pretrained YOLOV11l weights were fine tuned on a single class UXO despite ordnance type using Ultralytics [23]. Both training runs were configured for a maximum of 300 epochs using 1024 × 1024 pixel inputs, a batch size of 8, zero data loader workers, deterministic training, and an early stopping patience of 75 epochs. Optimizer selection was set to auto, and Ultralytics default parameters for AdamW with an initial learning rate of 0.002, a momentum parameter of 0.9, and a weight decay of 0.0005.

Ultralytics data augmentation was applied dynamically during YOLOV11l training. Mosaic augmentation was applied with a probability of 1.0 and disabled during the final 30 epochs. Horizontal flipping was applied with a probability of 0.50. Additional transformations included rotation of up to approximately +/- 5 degrees, translation of up to 10 percent of the image dimensions, random scaling using a scale parameter of 0.40, and HSV adjustments using hue, saturation, and value parameters of 0.015, 0.50, and 0.30, respectively. Vertical flipping, MixUp, CutMix, copy-paste, shear, and perspective transformations were disabled.

Ultralytics reported separate box-regression, classification, and distribution losses for the training and validation datasets during training. Validation was performed after each epoch, and Ultralytics retained the best.pt checkpoint. The 33 m model stopped after 263 epochs, with the best checkpoint identified at epoch 188, whereas the 15 m model completed all 300 configured epochs. The best checkpoint from each altitude regime was processed with the Ultralytics validation routine.

Ultralytics YOLOV11l validation used 1024 × 1024 pixel inputs, a minimum prediction confidence setting of 0.10, a non-maximum-suppression IoU setting of 0.50, and a maximum of 100 detections per image. Precision, recall, mAP50, and mAP50-95 were extracted from the Ultralytics results, and F1 was calculated as the harmonic mean of the reported precision and recall. The confidence setting controlled the minimum predictions retained for validation, while precision and recall were produced through the framework specific confidence curve calculations. These values were retained as default outputs of the YOLOV11l implementation and were interpreted as evidence of dataset utility rather than as fixed threshold counts directly equivalent to the RT-DETRR50 custom evaluator.

*RT-DETR-R50 Object Detection*

RT-DETR-R50 was used as a transformer based object detection model [24]. The models were initialized from the pretrained PekingU/rtdetr_r50vd_coco_o365 checkpoint [25] and fine tuned for single class UXO detection. Images were converted to RGB and processed using the Hugging Face RTDetrImageProcessor [25]. The 33 m model retained its original image processing configuration. To retain more spatial detail from the larger 15 m source images, the 15 m model used a fixed input size of 1600 × 1216 pixels. Both dimensions were divisible by 32 to maintain compatible feature map dimensions within the RT-DETR hybrid encoder. No explicit geometric or photometric data augmentation was implemented for RT-DETR-R50. This option was disregarded due to the more demanding computational requirements of the transformer based algorithm.

RT-DETR-R50 training was configured for a maximum of 50 epochs. The 33 m model used a batch size of 4, while the higher resolution 15 m model used a batch size of 1 with four gradient accumulation steps, retaining an effective batch size of 4. Both runs used zero data loader workers. AdamW optimization was applied with a learning rate of $2 \times 10^{-5}$ and a weight decay of $5 \times 10^{-4}$. The 15 m model was trained on Apple Metal Performance Shaders using bfloat16 autocast, while checkpoint evaluation was conducted in full precision, accelerating model training.

During 15 m training, mean training loss was recorded after every epoch. Full precision validation loss and detection metrics were calculated after epoch 1 and every two epochs thereafter. Validation calculated precision, recall, and F1 at confidence 0.10 and recorded confidence ranked average precision for checkpoint diagnostics. Checkpoint selection was based primarily on validation F1. When F1 values were tied within a tolerance of $10^{-4}$, standard mAP5095 was used as the first tie breaker and validation loss as the second tie breaker. Early stopping was configured after five evaluated epochs without improvement. Training completed all 50 epochs, and the best checkpoint was identified at epoch 46.

For the final RT-DETR-R50 evaluation, predictions with confidence scores below 0.10 were removed, and the remaining predictions were ranked by confidence. Predicted boxes were matched one to one with previously unmatched ground truth boxes using an IoU threshold of 0.50. Successful matches were counted as true positives, unmatched predictions as false positives, and unmatched annotations as false negatives. Precision, recall, and F1 were calculated from these counts. Average precision was calculated, with mAP50 calculated at IoU 0.50 and mAP50-95 calculated across IoU thresholds from 0.50 to 0.95 in increments of 0.05. Because predictions below confidence 0.10 were removed before the precision-recall curves were constructed, the RT-DETR-R50 AP values were conditioned on this confidence floor.

*Evaluation Scope and Interpretation*

YOLOV11l and RT-DETR-R50 were trained to determine how successful the labeled thermal datasets could support automated candidate detection. Each model retained its native training environment, checkpoint selection procedure, prediction postprocessing, and evaluation method. Readers of this work should understand that the study was not designed as a controlled experiment intended to benchmark or directly compare the two algorithms. Therefore, performance was interpreted within each model implementation and across flight altitudes. Because the native evaluators used different calculations and the models produced differently calibrated confidence scores, numerical differences between YOLOV11l and RT-DETR-R50. This approach reflects the objective of assessing dataset usability, establishing experimental proof-of-concept, and identifying practical considerations for future users rather than selecting a universally superior detector.

# Results

*Raw Image Collection and Labeled Dataset Outcomes*

The final raw source inventory from the four field campaigns contained 5,855 thermal image label pairs. Of these, 918 were positive images with at least one visible inert ordnance annotation and 4,937 were background images. The three 33 m collections contributed 2,626 images, while the June 15 m collection contributed 3,229 images. After resampling, the selected 33 m dataset contained 526 images and the selected 15 m dataset contained 786 images. The final validation subsets contained 106 images with 176 annotated instances at 33 m and 157 images with 192 annotated instances at 15 m.

*YOLOV11l Results*

The YOLOV11l results showed that the 15 m acquisition regime produced higher validation metrics than the 33 m regime. Precision increased from 0.890 to 0.925, recall increased from 0.828 to 0.833, and F1 increased from 0.858 to 0.877. The mAP50 increased from 0.865 to 0.905, while mAP50-95 increased from 0.488 to 0.664. The largest change occurred for mAP50-95, an increase of 0.176, indicating that the labeled 15 m dataset supported more consistent bounding box localization within the YOLOV11l model (see Table 3 and Figure 5).

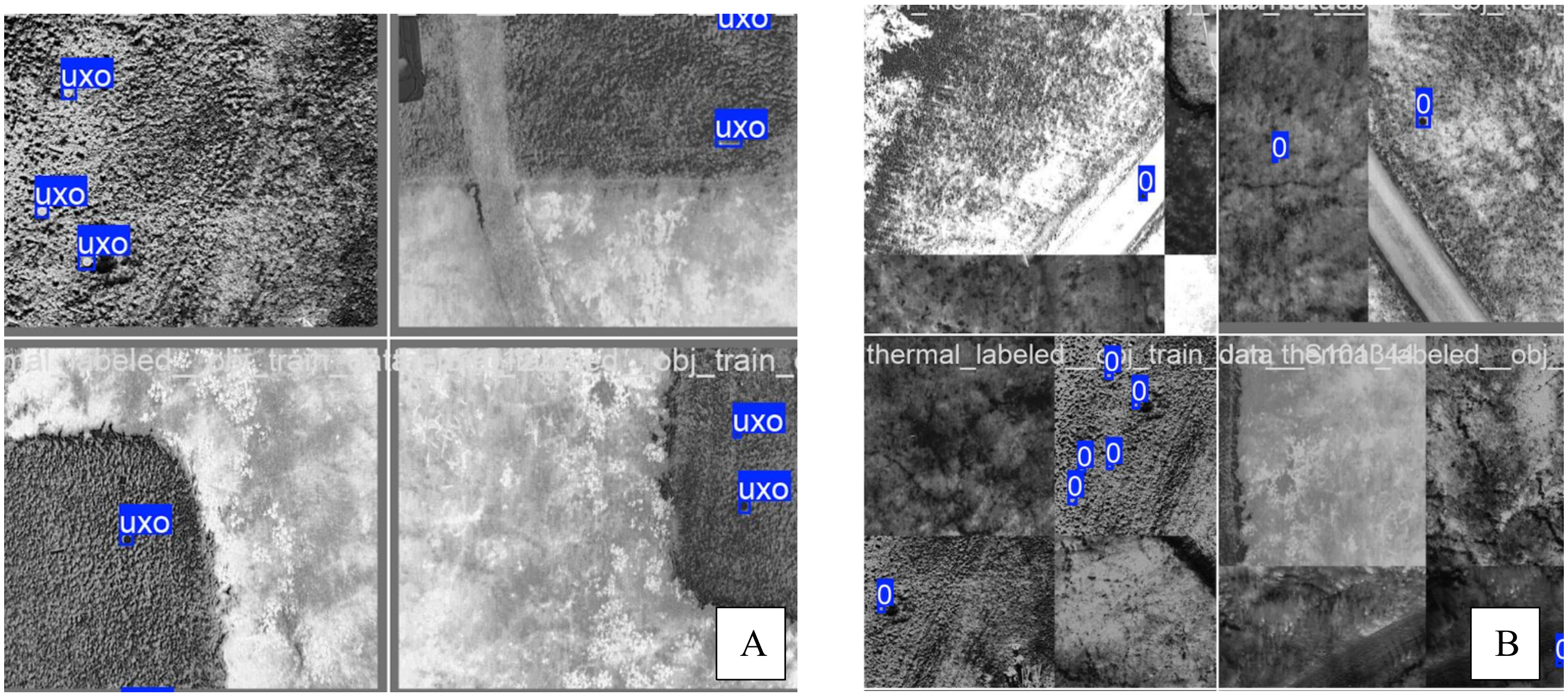

Figure 5A-B. Image (A) displays an example of the validation test output from Ultralytics. UXO are symbolized by blue box and "UXO" label. Image (B) represents the labeled training dataset, highlighted be the blue box and the class number "0".

*Table 3. Ultralytics validation results for YOLOV11l. The value 0.10 was the minimum prediction/retention threshold supplied to validation, and 0.50 was the non-maximum suppression IoU setting. Precision, recall, mAP50, and mAP50-95 are Ultralytics summary metrics; F1 was calculated as the harmonic mean of the reported precision and recall.*

| Regime | Model | Evaluation | Precision | Recall | F1 | mAP50 | mAP50-95 |
|---|---|---|---|---|---|---|---|
| 33 m | YOLOV11l | Ultralytics; min conf 0.10; NMS IoU 0.50 | 0.890 | 0.828 | 0.858 | 0.865 | 0.488 |
| 15 m | YOLOV11l | Ultralytics; min conf 0.10; NMS IoU 0.50 | 0.925 | 0.833 | 0.877 | 0.905 | 0.664 |

*RT-DETR-R50 Results*

The RT-DETR-R50 results also increased under the 15 m acquisition and preprocessing configuration. At 33 m, precision was 0.891, recall was 0.699, F1 was 0.783, mAP50 was 0.644, and mAP50-95 was 0.316. At 15 m, precision was 0.911, recall was 0.750, F1 was 0.823, mAP50 was 0.740, and mAP50-95 was 0.511. The 15 m model was selected at epoch 46 and evaluated on the same 157 image validation partition used for YOLOV11l. Relative to the 33 m model, the 15 m model increased precision by 0.020, recall by 0.051, F1 by 0.039, mAP50 by 0.097, and mAP50-95 by 0.195. The largest difference occurred for mAP50-95, indicating more consistent bounding box localization across stricter IoU thresholds within the 15 m RT-DETR-R50 implementation.

*Table 4. Validation results for RT-DETR-R50. Predictions below confidence 0.10 were removed, up to 100 remaining predictions per image were matched to ground truth at IoU 0.50, and AP was calculated from confidence ranked predictions retained at or above the 0.10 confidence floor.*

| Regime | Model | Evaluation | Precision | Recall | F1 | mAP50 | mAP50-95 |
|---|---|---|---|---|---|---|---|
| 33 m | RT-DETR-R50 | Custom; conf 0.10; match IoU 0.50 | 0.891 | 0.699 | 0.783 | 0.644 | 0.316 |
| 15 m | RT-DETR-R50 | Custom; conf 0.10; match IoU 0.50 | 0.911 | 0.750 | 0.823 | 0.740 | 0.511 |

The validation subsets were set to contain approximately 70 percent positive images and 30 percent background images. Therefore, precision should not be interpreted as an operational false alarm rate for a survey containing the natural and much larger number of background images. Results in Tables 3 and 4 describe performance across the complete acquisition, preprocessing, training, and evaluation configurations used for each model. The 33 m and 15 m RT-DETR-R50 models did not use identical input resolution policies, while YOLOV11l and RT-DETR-R50 retained different evaluators and differently calibrated confidence scores.

# Discussion

*Field Observations and Acquisition Timing*

Manual inspection of the thermal imagery suggested that targets were most distinguishable after changes in solar exposure or change in incidence angle. These periods generally best from directly after a change until approximately 30 to 60 minutes after sunrise, approximately 30 to 60 minutes after sunset, and immediately after cloud cover interrupted a prolonged period of direct sunlight. This behavior is related to differences in thermal inertia between the ordnance and the surrounding vegetation or surface material. For example, tall vegetation appeared to change temperature more quickly than short grass or bare ground. These changes also appeared to occur more rapidly during the summer campaigns. Therefore, these time periods should be considered preliminary scheduling guidance rather than statistically validated optimal collection windows.

*Detection Pipeline Performance and Acquisition Regimes*

Within the YOLOV11l pipeline, the 15 m acquisition regime produced higher values for every reported metric. The increase in mAP50-95 from 0.488 to 0.664 was the largest difference and suggests that bounding box localization was more consist
ent under the 15 m collection conditions. The greater number of pixels occupied by small targets at lower altitude may have contributed to this result. However, the two acquisition regimes also differed in UAV platform, thermal sensor, collection site, season, surface material, and source imagery. For this reason, the YOLOV11l results support the 15 m regime as a promising configuration for this implementation.

The RT-DETR-R50 results followed the same broad direction as the YOLOV11l results, with higher precision, recall, F1, mAP50, and mAP50-95 for the 15 m implementation. The largest difference occurred in mAP50-95, which increased from 0.316 at 33 m to 0.511 at 15 m. Input preprocessing was an important component of the 15 m implementation because its source images were substantially larger than the 33 m thermal frames. The fixed 1600 × 1216 pixel input was selected to retain more of the available small target detail while maintaining compatible RT-DETR feature map dimensions.

It is important to consider the balance between survey coverage and target representation when planning an operational mission. Higher flight altitudes can increase area coverage and achieve a reduced collection time, while lower altitudes can increase the number of pixels representing a small target. In all reality, altitude choice would depend on an array of factors, most especially the user. A demining team in a post conflict AOI would conceivably have more time to conduct slower flights/imagery collection at low altitudes than an EOD unit or demining crew in or very close to an active conflict zone.

From an operational perspective, a practitioner would implement one complete pipeline rather than conduct an architecture comparison during a field survey unless an array of computation resources are available. Algorithm selection may depend on available compute, software familiarity, model support, processing speed, and the desired balance between missed detections and false alarms. Each selected model should be calibrated and independently validated with representative site data (i.e., ordnance type and spatial distribution) before use. Accordingly, our results demonstrate two alternative routes through the end to end methodology rather than a benchmark establishing that one detector is better than the other. That being said, deployment of a pretrained object detection model, such as presented in this work, would most likely be efficient at providing a first look opportunity during initial UXO surveys. It should also be mentioned, despite the fact that both of these tested algorithms displayed successful results, Yolov11 caries much lighter computational load and generally trains more quickly than RT-DETR and other transformer models as of July 2026.

*Operational Decision Support*

After initial surveys and fine tuning, a potential UXO map would be used to prioritize technical survey, EOD assessment, or another approved form of ground investigation, but it would not indicate that an area was free of explosive hazards. Future studies should evaluate this performance in addition to image level detection metrics by reporting the percentage of physical targets detected, false candidate locations per area or flight, localization error, processing time, and analyst review workload. These measures would provide more operationally meaningful information than image level precision, recall, F1, and mAP alone. Blind candidate map evaluation was not completed in this study and remains an important next step in our goals toward operational validation.

*Recommendations for Humanitarian Mine Action and Demining*

Several practical considerations emerged from the field collection, labeling, and object detection evaluations. Thermal and RGB imagery should be collected together so that thermal anomalies can be interpreted against visible surface conditions. This practice will aid in giving the user more contextual information of a scene. For broad detection campaigns, training data should include varied substrates, vegetation heights, target sizes, target orientations, and substantial background only imagery because these factors can influence thermal contrast and false detections. Most importantly, collection schedules should consider periods following changes in solar exposure, including sunrise, sunset, and cloud transitions, while recognizing that the timing observations in this study were opportunistic. Flight altitude should be selected by balancing area coverage against versus temporal need, and especially, the safety of demining teams. Finally, any selected detector should be calibrated using representative local imagery and used to prioritize qualified analyst review rather than to make an independent clearance decision (see Figure 6A-B).

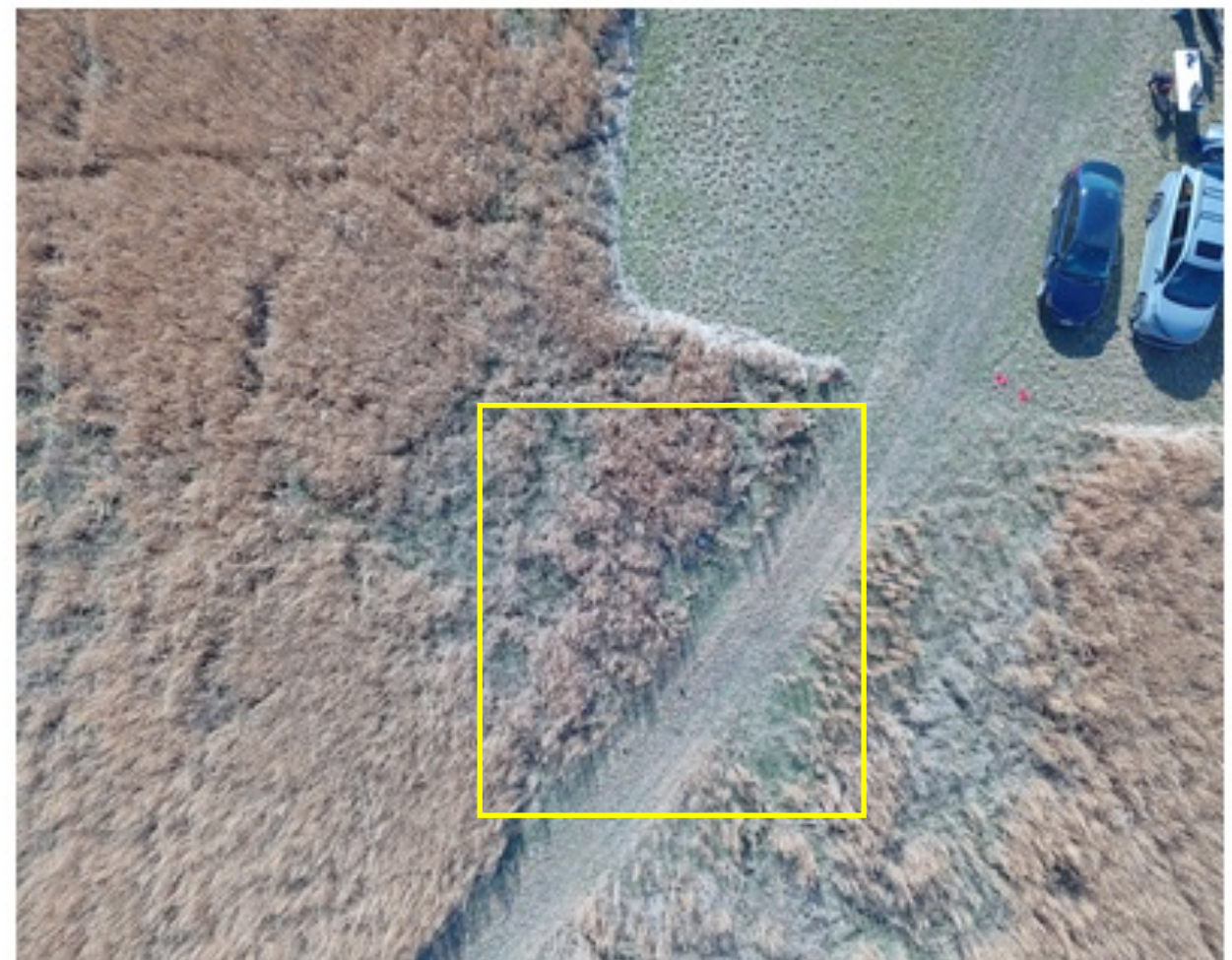

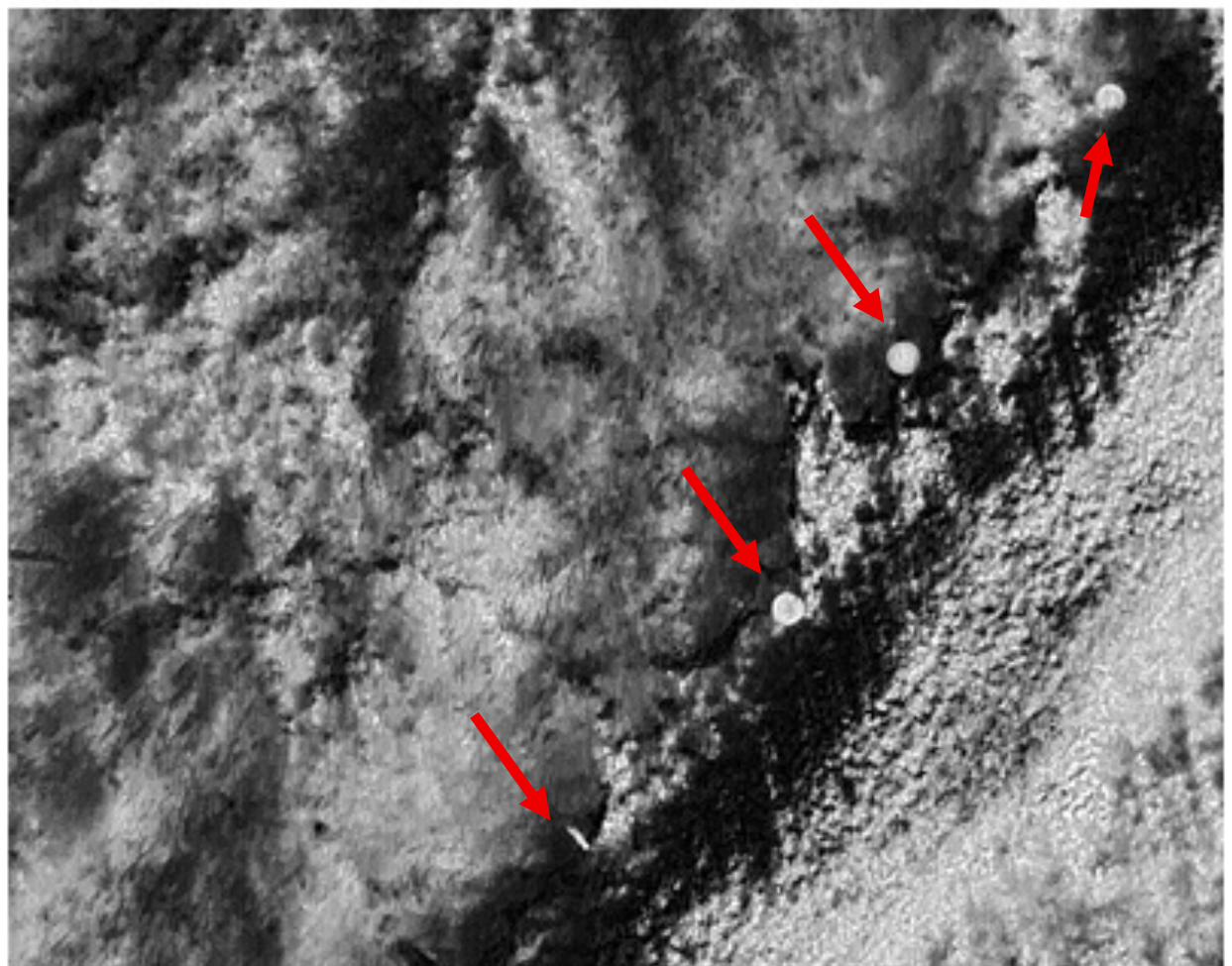

*Figure 6A B. Comparison of RGB and thermal imagery collected from the same scene near dusk. Several inert items are obscured by vegetation in the RGB image (A) but remain visible as bright circular thermal anomalies in the thermal image (B). The yellow box in image (A) represents FOV of image (B). Red arrows in image (B) point to inert objects visible in thermal.*

*Limitations and Operational Boundaries*

All items used in this study were inspected by EOD specialists and verified to be inert. Readers of this work should consider that the removal or absence of explosive material altered the mass, density, heat capacity, and internal heat transfer of the items. Therefore, the observed thermal behavior cannot be assumed to completely represent live ordnance. In addition, the collection included materials such as steel, aluminum, Bakelite, rubber, and plastic. As expected, these differences in material and construction produced a range of thermal responses.

Targets were deliberately placed, and many items in tall vegetation were marked with red survey flags. It is possible that thermal shadows from disturbed vegetation during the placement of flags could have provided contextual information.
The Norris campaigns were collected with 80 to 85 percent image overlap, and the final dataset audit confirmed that no exact image hashes were shared between training and validation. However, the partitions were still created at the image level. Neighboring frames may show the same physical item under nearly identical conditions even when the files are not exact duplicates. Therefore, similar sections of frames could conceivably remain in both subsets, creating a risk that the validation results are more optimistic than performance on completely independent flights.

Background images were randomly reduced so that approximately 70 percent of the selected images were positive. This step was useful for model development, but it does not represent a potentially very low target prevalence expected during an operational survey, nor the alternative. For this reason, precision calculated from the altered validation sets should not be interpreted as the percentage of correct detections expected over an AOI. Future studies might include complete background flights and report false detections in an AOI per flight. That being said, not every minefield or conflict zone will have the same distribution of ordnance placement or ordnance type. It is imperative that users of such technologies be aware of tactical use of mines and other projectiles by parties to the conflict. For example, a conventional army will use mines and ordnance typically in large swaths to hinder or direct movement where as an insurgency might place mines or other explosives near transportation hubs or choke points.

The 15 m and 33 m acquisition campaigns differed in platform, season, and dataset composition. The final 15 m RT-DETR-R50 model was trained using the same source image pool and train validation partitions as YOLOV11l, and our independent audit found no exact image content duplicates between the training and validation sets. The 15 m RT-DETR-R50 model used fixed 1600 × 1216 pixel inputs to retain more detail from the larger source images, whereas the 33 m implementation retained its original preprocessing configuration. The model implementations also differed in data augmentation, training duration, and evaluation procedures. Consequently, the reported performance metrics should not be interpreted as isolating the effect of altitude or model architecture. Given these methodological differences, together with the inherent variability and confounding characteristics of the imagery collection campaigns, the relative performance of the models and data collection altitudes cannot be directly attributed to one factor.
Finally, this study evaluated controlled placements of surface and partially obscured inert ordnance rather than unknown live hazards or independently buried targets. The methodology should be considered a screening and decision support approach until it has been evaluated with independent flights, geospatial grouping of detections, target level scoring, and approved operational field procedures.

## Conclusions

*Dataset Development, Object Detection, and Recommended Use*

This study created a multi campaign UAV thermal image data set, converted the raw imagery into labeled single class object detection datasets, and trained and evaluated YOLOV11l and RT-DETR-R50 to test automated inert ordnance screening. Across four controlled field campaigns, the final source inventory contained 5,855 thermal image label pairs, including 918 positive images and 4,937 background images. The strongest YOLOV11l results were obtained with the 15 m dataset, which produced precision of 0.925, recall of 0.833, F1 of 0.877, mAP50 of 0.905, and mAP50-95 of 0.664. The strongest RT-DETR-R50 results were also obtained with the 15 m dataset, which produced precision of 0.911, recall of 0.750, F1 of 0.823, mAP50 of 0.740, and mAP50-95 of 0.511. These findings demonstrate that both labeled datasets supported candidate detection under controlled conditions and that preserving input detail was important for the larger 15 m imagery.

Practical recommendations from the study include collecting thermal and RGB imagery together; incorporating varied vegetation, surfaces, target sizes, orientations, and background only imagery; considering periods after changes in solar exposure; selecting altitude by balancing coverage against target representation; and validating the selected model with representative local data. Model outputs should be used to prioritize analyst review and follow on technical survey or EOD assessment, not to declare an area free of explosive hazards. Future work should include geospatial grouping, target level scoring, threshold calibration, and approved operational field testing. The study therefore provides a dataset development and screening framework that can be expanded as these additional steps are completed.

## Acknowledgments

This work would not have been possible without the technical assistance, field and flight collection support, and access to inert training materials provided by the leadership and EOD specialists of the U.S. Army 52nd EOD Group at Fort Campbell, Kentucky; retired Sergeant Major Mike Vining; Berlin Seaborn; the Tennessee National Guard; Knoxville Tactical; Crossroads Firearms; the John Sevier Hunter Education Center; L&M Landscape Supply, LLC; Jay Miller; and Gerald and Lois Crawford.

*Author Contributions*

Chad Melton: Concept and experimental design, data collection, data labeling, model training, and article composition

Annabelle Kelton: Data collections, and data labeling, editing.

*Funding*

This work was funded by UT-Battelle/Oakridge National Laboratory

Data Availability Statement

*Conflicts of Interest*

The authors declare no conflict of interest.